\pdfoutput=1
\documentclass[11pt,a4paper]{article}

\usepackage[T1]{fontenc}
\usepackage[utf8]{inputenc}
\usepackage[english]{babel}
\usepackage{lmodern}
\usepackage[margin=2.5cm]{geometry}
\usepackage{microtype}
\usepackage{graphicx}
\usepackage{tabularx}
\usepackage{booktabs}
\usepackage[numbers,sort&compress]{natbib}
\usepackage[hidelinks]{hyperref}
\usepackage{xcolor}

\title{Evaluating Large Language Models for Scientific Paper Search: Retrieval Success and Inference Latency on Mac Studio and NVIDIA A100}
\author{Frederic Andriantseheno Rakoto\\ \normalsize Mnemosyne\\ \normalsize Inria Bordeaux, France
\and Yannis Bendi-Ouis\\ \normalsize Mnemosyne\\ \normalsize Inria Bordeaux, France
\and Xavier Hinaut\\ \normalsize Mnemosyne\\ \normalsize Inria Bordeaux, France}
\date{}

\begin{document}
\maketitle

\begin{abstract}
\noindent Scientific literature search can benefit from locally deployed LLM assistants, which reduce reliance on hosted model providers while giving research teams greater control over their data and infrastructure. This work investigates whether open-weight models can support web-assisted paper retrieval under practical computing constraints. We first characterize inference latency as the input context grows from 512 to 128k tokens on a Mac Studio and a server equipped with two NVIDIA A100 GPUs. We then evaluate eight models on the retrieval of 70 papers from seven scientific repositories, using queries that represent three levels of domain knowledge. The results show that retrieval success depends strongly on the selected model and the specificity of the query. And that A100 GPUs are generally faster than the Mac Studio, but its unified memory allows larger models that could not be loaded in the tested GPU configuration. 
\end{abstract}

\section{Introduction}

Scientific assistants may process unpublished manuscripts, hypotheses, source code, and experimental data. Sending this material to an external language-model provider creates confidentiality, governance, and dependency concerns. These concerns depend on the service and its settings. Running models within the laboratory keeps model inputs under institutional control and reduces dependence on remote APIs. Local LLM alone does not produce an effective scientific agent. Agent behavior also depends on its \emph{harness}, the software layer that constructs context, manages memory, exposes tools, schedules model calls, and verifies their results. Recent work treats this layer as a major component of agent performance \citep{gu2026harness}. Evaluations with fixed runtime models also show that changing the harness can substantially alter task success and execution cost \citep{wu2026harnessdev}. A research laboratory may not have the resources to train models comparable to frontier proprietary systems, but it can improve, inspect, and control the system built around an open-weight model. \\

\noindent The open-source agent ecosystem has expanded rapidly. At the beginning of this project, however, the available solutions did not provide a stable fit for the laboratory's requirements, including local inference, literature-search tools, explicit context control, and maintainability. We therefore developed our own harness to find and work with scientific papers.
\noindent This paper evaluates whether locally deployed open-weight models provide a practical foundation for this use case. We measure TTFT and total response time from 512 to 128k input tokens on a Mac Studio and an NVIDIA A100 configuration. We also evaluate eight models on a web-assisted task involving 70 scientific papers, seven publication sources, and three levels of prompt specificity. We ask how latency varies across deployments and context lengths, how retrieval success varies across models and prompts, and which trade-offs support a local scientific-literature agent.

\section{Related Work}

LLM deployment depends on the serving engine as well as the hardware and model. \emph{vLLM} is a widely adopted serving engine for LLMs, providing parallel request and multi-GPU execution on NVIDIA CUDA GPUs \citep{kwon2023pagedattention}. 
\emph{llama.cpp} is a lightweight serving engine designed for efficient local execution with low resource, and supports a wide range of quantization schemes through the GGUF model format \citep{ggerganov2026llamacpp}. On Apple Silicon, \emph{MLX} is designed around unified memory, allowing CPU and GPU operations to share arrays without explicit transfers \citep{hannun2023mlx}. Several quantization methods are available for LLM inference, including \emph{AWQ} and \emph{GPTQ}. \\

\noindent The evaluated models come from the Qwen, GLM, and Gemma families. We selected both dense and Mixture-of-Experts (MoE) designs \citep{yang2025qwen3} \citep{glmteam2025glm45} \citep{gemmateam2026gemma4}. In an MoE model, a router activates only a subset of experts for each token. This separates total parameter count from active parameter count and can reduce computation, although the complete set of weights must still be stored usually in VRAM \citep{fedus2022switch}. This distinction is relevant to the A3B, A4B, and A10B models in our experiments. \\

\noindent An agent's capabilities also depend on its harness, which assembles context, exposes tools, manages state, and controls the model/tool loop. Agent loop implementation established a common pattern in which reasoning steps alternate with external actions \citep{yao2023react}. Recent open-source harnesses implement this pattern with different priorities. \emph{OpenClaw} provides a self-hosted assistant through a gateway that connects models, tools, automations, and messaging channels \citep{openclaw2026}. \emph{Hermes Agent} focuses on persistent memory, reusable skills, and learning across sessions \citep{nousresearch2026hermes}. \emph{Pi} separates a model API, an agent loop, and a coding interface into reusable components \citep{earendil2026pi}. \emph{Qwen Code} instead centers the developer workflow through terminal, editor, and software-development integrations while retaining support for multiple model providers \citep{qwen2026code}. \emph{DeepSeek Harness} adopts an “everything is a plugin” architecture to maximize component interchangeability \citep{deepseek2026harness}. Even thought both these systems allows different models, some models have been trained with a specific harness, making it better with this one.\\

\noindent Scientific search introduces a more specific evaluation problem. \emph{LitSearch} evaluates retrieval and reranking on 597 queries about machine-learning and natural-language-processing papers \citep{ajith2024litsearch}. \emph{PaperQA2} combines search, evidence gathering, and citation traversal to answer scientific questions \citep{skarlinski2024paperqa2}. Our study focuses on whether locally executed models using the same web-search tool can identify a requested paper from prompts of increasing expertise. It combines this evaluation with latency measurements on Mac Studio and Nvidia A100.

\section{Materials and Methods}

\subsection{Platforms}

The first local platform is a M3 ultra Mac Studio with 512 GB of unified memory. Its shared CPU–GPU memory pool (called unified memory) can hold models larger than the memory of a single NVIDIA GPU, but the configuration does not offer parallel request. Apple reports 9 W at idle and 270 W under maximum load for an M3 Ultra Mac Studio\citep{apple2025macstudiopower}.
The second local platform provides one compute node with two NVIDIA A100 GPUs, each with 40 GB of VRAM (HBM2), and 512 GB of RAM\citep{plafrim2026hardware}. The two accelerators allow parallelism request, although the configuration does not benefit from NVLink for inter-GPU communication. A single NVIDIA A100 GPU can have a maximum thermal design power of up to 400 W \citep{nvidia2021a100}. LM Studio serves GGUF models through llama.cpp on the Mac, whereas vLLM serves AWQ quantized models on NVIDIA A100. The comparison therefore covers complete deployment configurations, combining hardware, runtime, and weight format.

\subsection{Models}

The evaluation covers 4 Qwen variants, Qwen3.5-27B, Qwen3.5-35B-A3B, Qwen3.5-AgentWorld-35B-A3B, and Qwen3.5-122B-A10B, together with Gemma-4-31B, Gemma-4-26B-A4B, GLM-4.7-355B-A32B and GLM-4.7-Flash-30B-A3B. The A3B, A4B, A10B and A32B suffixes denote active parameters in the named MoE variants.
Qwen3.5-122B-A10B and GLM-4.7-355B-A32B exceeded the available NVIDIA A100 GPU memory, leaving six models measured on both platforms. \\

\begin{figure}[htbp]
\centering
\includegraphics[width=0.9\linewidth]{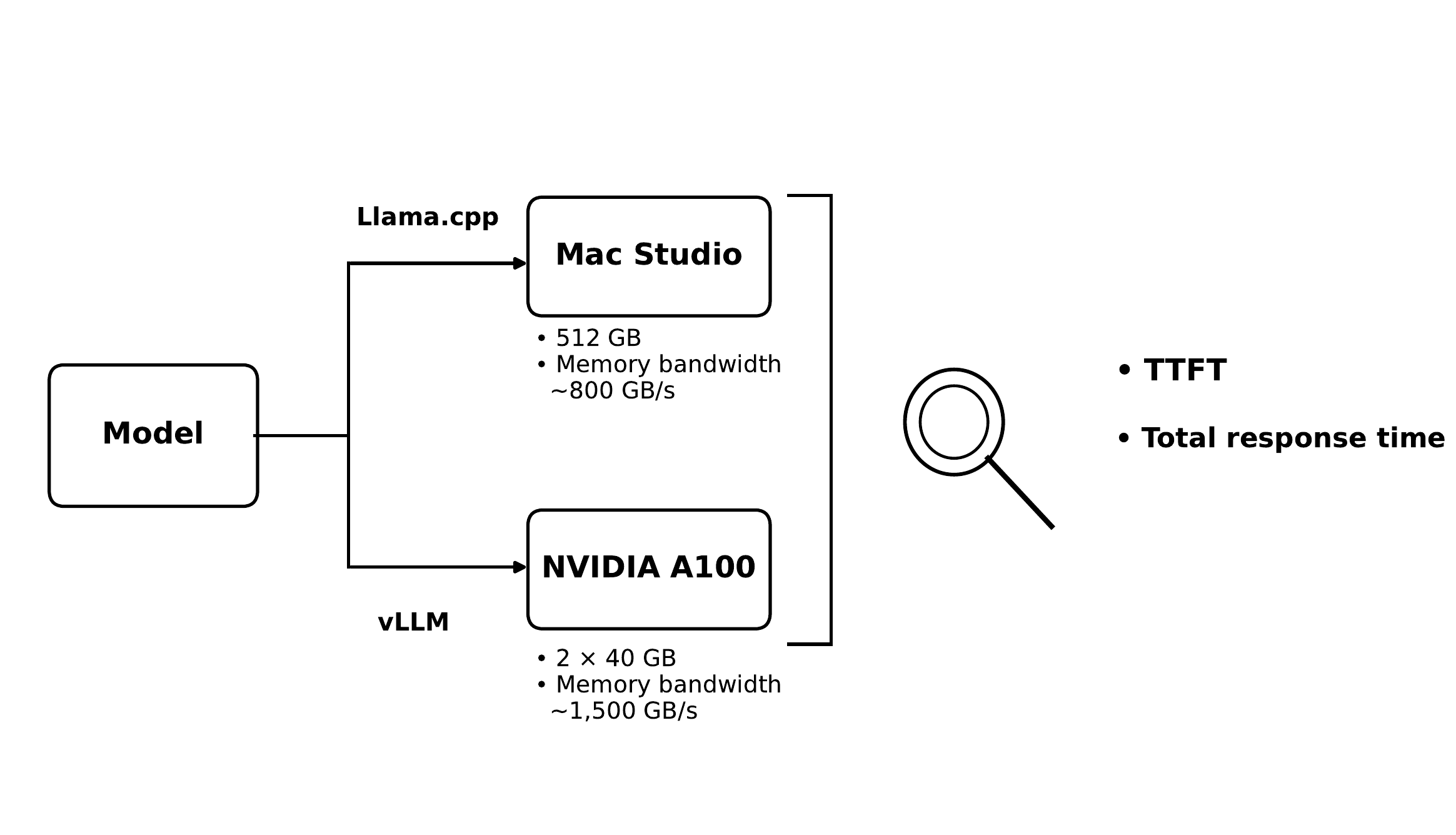}
\caption{Inference-latency benchmark workflow. Models are served through LM Studio on the Mac Studio and vLLM on two NVIDIA A100 40 GB GPUs. TTFT and total response time are recorded for each request.}
\label{fig:workflow}
\end{figure}

\noindent The figure \ref{fig:workflow} illustrates the benchmark workflow: each model is served on one or both platforms if it fits both, and the client records time to first token (TTFT) and total response time for each input length.

\subsection{Inference latency}

The client sends one user message and limits the response to 100 tokens. Input lengths are 512, 1k, 2k, 4k, 16k, 32k, 48k, 64k, 96k, and 128k tokens, with k = 1,024. For each model, the string «\texttt{lol }» was repeated and adjusted until the server-reported token count reached each target.
For every request, the client starts a timer immediately before sending the HTTP request. Streaming is enabled. The elapsed time recorded when the first token is received is the TTFT. The client then consumes the complete stream and records the elapsed time again when the response ends. This second value is the total response time. \\

\noindent After loading each model, a preliminary request «\texttt{test}» is sent to warm the serving environment. This request is excluded from the measurements. The ten requests are then processed once and sequentially.\footnote{Despite repeated attempts, TTFT measurements for Gemma-4-26B-A4B on the Mac could not be obtained at 96k and 128k tokens due to a model-specific issue that could not be resolved within our experimental setup. These records are therefore excluded from complete-timing comparisons.}

\subsection{Scientific paper search}

\begin{figure}[htbp]
\centering
\includegraphics[width=0.9\linewidth]{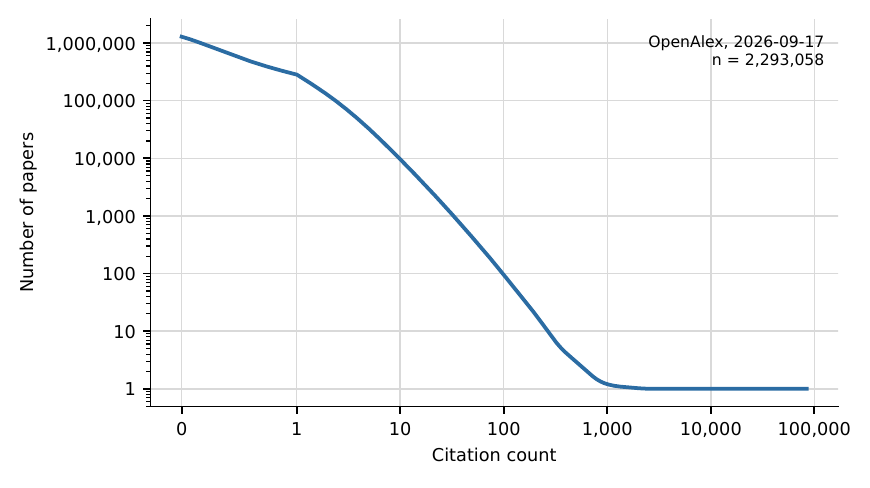}
\caption{Smoothed distribution of arXiv papers by citation count according to OpenAlex \citep{priem2022openalex}. The curve shows that papers with few citations are much more common than highly cited papers. }
\label{fig:paper_counts}
\end{figure}

The corpus contains 70 targets, ten each from \emph{arXiv}, \emph{bioRxiv}, \emph{medRxiv}, \emph{ChemRxiv}, \emph{SSRN}, \emph{Research Square}, and \emph{agriRxiv}. Because few papers are highly cited, as shown in figure \ref{fig:paper_counts}, paper selection combined random sampling with variation in citation counts to include both low and highly cited publications.The code used for dataset construction and the experiments is publicly available in the FredAgent repository.\footnote{\url{https://github.com/AndriantsehenoFrederic/FredAgent/}}

\noindent Before evaluation, Google Gemini 3.5 Pro generated three search queries for each target. The \texttt{basic} level expresses the topic in accessible terms, the \texttt{intermediate} level introduces discipline-specific concepts and methods, and the \texttt{expert} level combines distinctive technical clues. The resulting 210 queries were fixed before testing and reused unchanged for every evaluated model. The following example shows the three queries used for the paper \emph{The Symbol Grounding Problem}. The target title itself was not included in the request sent to the evaluated model.

\begin{center}
\small
\begin{tabularx}{\linewidth}{@{}>{\raggedright\arraybackslash}X>{\raggedright\arraybackslash}X@{}}
\toprule
\textbf{Information level} & \textbf{Query sent to the evaluated model} \\
\midrule
\texttt{basic} & how symbols and words can acquire meaning from perception instead of only referring to other symbols \\
\texttt{intermediate} & grounding formal symbols in sensory iconic representations and learned categorical feature detectors in cognitive models \\
\texttt{expert} & hybrid connectionist symbolic cognition bottom-up grounding category names proximal sensory projections invariant features Chinese dictionary analogy \\
\bottomrule
\end{tabularx}
\end{center}

\noindent All eight models were tested on 70 targets with three query levels, producing 1,680 evaluations. Search success is the number of positive labels divided by the number of evaluations. The overall score for a model is it overall score of success on those 210 queries.

\section{Results}

\subsection{Inference latency}

\begin{figure}[htbp]
\centering
\includegraphics[width=0.9\linewidth]{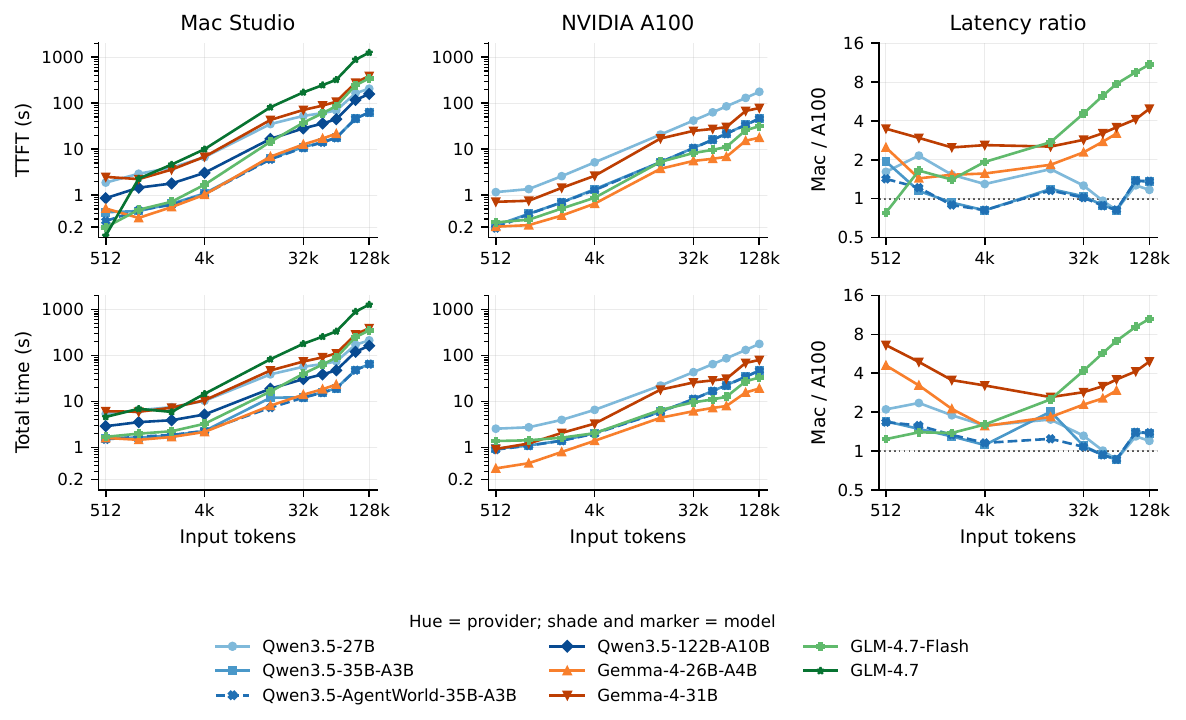}
\caption{TTFT (top) and total response time (bottom) versus input length on the Mac Studio and A100. The latency ratio is defined as Mac duration divided by A100 duration. Colors identify providers, while markers distinguish their models.}
\label{fig:res-inf}
\end{figure}

As shown in figure \ref{fig:res-inf}, latency increases with input length on both platforms, with the sharpest rise occurring on the Mac beyond 64k tokens. When the input grows from 64k to 96k, a factor of 1.5, Mac TTFT increases by factors of 2.4–2.8 across the seven models with complete measurements. At 128k, TTFT represents 92.7–99.3\% of total response time for 100 output tokens generated. 

\noindent On the Mac, Qwen3.5-AgentWorld-35B-A3B and Qwen3.5-35B-A3B remain the fastest complete configurations. Their total response times are approximately 19 seconds at 64k and 65 seconds at 128k. Gemma-4-26B-A4B belongs to the same leading group through 64k, where it takes 23 seconds, even though its missing TTFT values prevent a complete comparison at longer inputs. The larger Qwen3.5-122B-A10B can run on the Mac, although its 162 seconds total time at 128k is more than twice that of the two leading configurations. The ordering changes on the A100. Gemma-4-26B-A4B is the fastest model at 128k with a total time of 19 seconds, followed by GLM-4.7-Flash at 33 seconds. The two Qwen A3B variants follow at approximately 47 seconds. Thus, the models leading on one platform are not necessarily those leading on the other. The A100 advantage is also model dependent. At 128k, its total time gain ranges from 1.2 to 10.6 across common models. The gain is approximately 1.4 for the two fastest Qwen configurations, but reaches 10.6 for GLM-4.7-Flash.

\subsection{Scientific paper search}

\begin{figure}[htbp]
\centering
\includegraphics[width=0.9\linewidth]{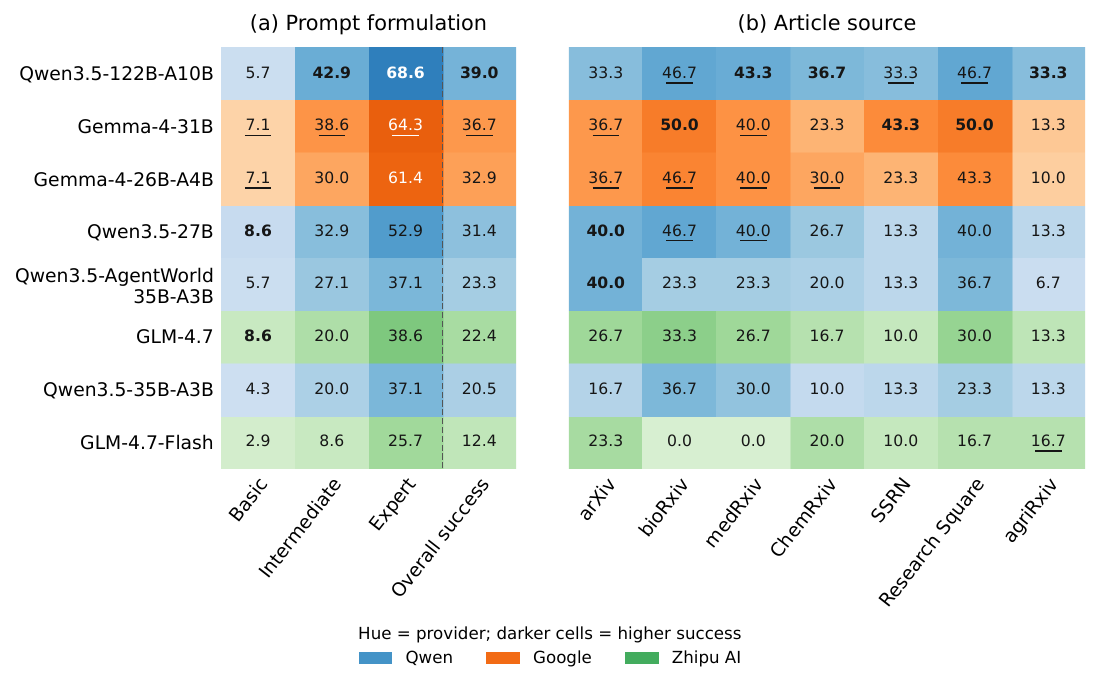}
\caption{Recorded search success rate for eight models. Color identifies the provider and color intensity represents success rate. The highest value in each column is bold and the second highest distinct value is underlined, with ties formatted identically. Each model has 70 evaluations per query-level cell, 210 for overall success, and 30 per source.}
\label{fig:res-details}
\end{figure}

Figure \ref{fig:res-details} summarizes the 1,680 search evaluations (3 queries $\times$ 70 papers $\times$ 8 models). Prompt specificity produces the clearest overall pattern. Across the eight models, success increases from 6.2\% for \texttt{basic} queries to 27.5\% for \texttt{intermediate} queries and 48.2\% for \texttt{expert} queries. Every model follows this progression, indicating that distinctive technical clues substantially improve retrieval.

\noindent Three models form the leading group. Qwen3.5-122B-A10B obtains the highest overall success rate at 39.0\% and reaches 68.6\% with expert queries. Gemma-4-31B and Gemma-4-26B-A4B follow with overall rates of 36.7\% and 32.9\%, respectively, and both exceed 60\% with the expert queries. The remaining models achieve at most 31.4\% overall, so model choice remains important even when every model receives the same queries and search tool.
Success also varies with publication source. Research Square and bioRxiv obtain the highest aggregated rates, both close to 36\%, while agriRxiv is lowest at 15.0\%. These values are descriptive rather than a ranking of scientific fields because each source contains only ten target papers.

\section{Discussion and Conclusion}

Taken together, the experiments support local LLM execution as a practical basis for web-assisted scientific paper search when results remain subject to human verification. No model is optimal for every use. Qwen3.5-122B-A10B is the most promising option when retrieval is the priority, although its higher latency makes it less suitable for highly interactive work. Gemma-4-26B-A4B provides the most balanced combination of retrieval success and responsiveness among the evaluated models even so we missed TTFT for 96k and 128k input token. The Qwen A3B variants favor speed over retrieval success and are more appropriate when rapid successive calls matter. These conclusions remain conditional on validation of the recorded retrieval labels. \\

\noindent For a small research team, the Mac Studio is a practical local platform. LM Studio simplifies model installation and replacement, local inference keeps prompts and unpublished material within the laboratory, and unified memory supports experimentation with models that do not fit in the tested GPU configuration. The indicative acquisition estimate is EUR 7,000–8,000 for the Mac, compared with EUR 22,000–26,000 for two A100 40 GB cards alone. A complete A100 server would require additional components, including CPUs, RAM, storage, power supplies, cooling, and the server chassis, potentially increasing the total acquisition cost substantially. These figures do not include energy, maintenance or utilization. We can also note that in terms of privacy web-assisted retrieval also continues to expose search queries to external services.
A single Mac can nevertheless become a bottleneck when several researchers submit long or repeated requests. Our sequential protocol does not measure concurrent throughput \cite{bendi2024deploying}, but a server with NVIDIA GPUs and vLLM offers more opportunities for request and model parallelism. Such a platform is therefore preferable for intensive multi-user service or agents that perform many successive model calls. Coding agents are one example because interaction speed often matters throughout the development loop. On the Mac, models with few active parameters provide an alternative for this type of workload, although coding quality was not evaluated here. \\

\noindent The Mac's 512 GB unified-memory capacity allowed Qwen3.5-122B-A10B and GLM-4.7 to run locally, whereas neither fit in the tested pair of 40 GB A100 GPUs. More GPUs, offloading, or stronger quantization could change this limit but each came with a cost or a tradeoff. Quantization reduces memory requirements but may alter model behavior and output quality, even though AWQ is designed to limit accuracy loss \citep{lin2024awq}. Retrieval was evaluated only on the Mac because the experiment targets model capability rather than hardware speed. Identical model representations should give comparable performance across machines, even though different quantization formats may produce different outputs. \\

\noindent For the intended use of FredAgent, these results led us to retain the Mac Studio as the primary deployment platform. FredAgent is designed as a local scientific assistant for a small research team, where the ability to run a broad range of models, preserve control over sensitive research data, and maintain a simple local infrastructure is more important than minimizing inference latency alone. The Mac Studio’s large unified memory allows models to be tested and deployed that could not fit within the available A100 configuration, while its performance remains sufficient for the interactive literature-search workflow considered here. Although an NVIDIA GPU server remains advantageous for high-throughput or concurrent workloads, these benefits do not currently outweigh the additional infrastructure requirements for our use case. FredAgent will therefore rely on local inference on the Mac Studio, with model selection adapted to the task and retrieved scientific references remaining subject to human verification.

\bibliographystyle{unsrtnat}
\bibliography{references}

\end{document}